\documentclass[sigconf]{acmart}

\AtBeginDocument{%
  \providecommand\BibTeX{{%
    \normalfont B\kern-0.5em{\scshape i\kern-0.25em b}\kern-0.8em\TeX}}}

\setcopyright{acmcopyright}
\copyrightyear{2020}
\acmYear{2020}
\setcopyright{acmcopyright}\acmConference[MM '20]{Proceedings of the 28th ACM International Conference on Multimedia}{October 12--16, 2020}{Seattle, WA, USA}
\acmBooktitle{Proceedings of the 28th ACM International Conference on Multimedia (MM '20), October 12--16, 2020, Seattle, WA, USA}
\acmPrice{15.00}
\acmDOI{10.1145/3394171.3413837}
\acmISBN{978-1-4503-7988-5/20/10}

\acmSubmissionID{783}

\begin{document}
\fancyhead{}

\title{PIDNet: An Efficient Network for Dynamic Pedestrian Intrusion Detection}


\author{Jingchen Sun}
\affiliation{%
  \institution{State Key lab. of Industrial Control Technology, Zhejiang University, Hangzhou, China}}
\email{jingchensun@zju.edu.cn}

\author{Jiming Chen}
\affiliation{%
  \institution{State Key lab. of Industrial Control Technology, Zhejiang University, Hangzhou, China}
  \institution{Alibaba-ZJU Joint Research Institute of Frontier Technologies}
  }
\email{cjm@zju.edu.cn}

\author{Tao Chen}
\authornote{Corresponding Author.}
\affiliation{%
  \institution{School of Information Science and Technology, Fudan University, Shanghai, China}}
\email{eetchen@fudan.edu.cn}

\author{Jiayuan Fan}
\affiliation{%
  \institution{Academy for Engineering and Technology, Fudan University, Shanghai, China}}
\email{jyfan@fudan.edu.cn}

\author{Shibo He}
\affiliation{%
  \institution{State Key lab. of Industrial Control Technology, Zhejiang University, Hangzhou, China}
  \institution{Alibaba-ZJU Joint Research Institute of Frontier Technologies}
  }
\email{s18he@zju.edu.cn}

\renewcommand{\shortauthors}{Sun and Chen, et al.}

\begin{abstract}
Vision-based dynamic pedestrian intrusion detection (PID), judging whether pedestrians intrude an area-of-interest (AoI) by a moving camera, is an important task in mobile surveillance. The dynamically changing AoIs and a number of pedestrians in video frames increase the difficulty and computational complexity of determining whether pedestrians intrude the AoI, which makes previous algorithms incapable of this task. In this paper, we propose a novel and efficient multi-task deep neural network, PIDNet, to solve this problem. PIDNet is mainly designed by considering two factors: accurately segmenting the dynamically changing AoIs from a video frame captured by the moving camera and quickly detecting pedestrians from the generated AoI-contained areas. Three efficient network designs are proposed and incorporated into PIDNet to reduce the computational complexity: 1) a special PID task backbone for feature sharing, 2) a feature cropping module for feature cropping, and 3) a lighter detection branch network for feature compression. In addition, considering there are no public datasets and benchmarks in this field, we establish a benchmark dataset to evaluate the proposed network and give the corresponding evaluation metrics for the first time. Experimental results show that PIDNet can achieve 67.1\% PID accuracy and 9.6 fps inference speed on the proposed dataset, which serves as a good baseline for the future vision-based dynamic PID study.
\end{abstract}



\maketitle
\section{INTRODUCTION}
Nowadays, vision-based Pedestrian Intrusion Detection (PID) is becoming increasingly important for many emerging surveillance applications, especially in autonomous driving and intelligent transportation management. It determines whether a pedestrian exists in a restricted area of interest (AoI), and usually consists of two steps: detecting pedestrians in the image and then judging whether the detected pedestrians are located within the AoI. According to whether the surveillance camera is moving, the vision-based PID can be categorized into dynamic PID and static PID. In static PID, the surveillance camera is fixed, most home surveillance scenarios by a fixed surveillance camera belong to this case. While in dynamic PID, the camera may change from time to time, resulting in consistently changing frames. Take the autonomous driving as an example, when the car is moving, the front road area (considered as AoI here) and pedestrians captured by the camera in the car keeps changing and it is important to monitor whether a pedestrian suddenly enters into the AoI. Compared with the static PID problem, the dynamic PID is clearly more difficult to address.

Previous studies on vision-based PID mainly focused on static PID, and adopted handcrafted features based \cite{liang2012real,zhang2015perimeter,matern2013automated} methods such as optical flow and frame difference for intruder detection. Prabhakar et al. \cite{prabhakar2012} and Wang et al. \cite{video5} exploited the background subtraction to detect intrusion behavior. Zhang et al. \cite{video2} proposed an algorithm based on GMM (Gaussian Mixture Model) and SVM (Support Vector Machine) to identify the image foreground and classify the pedestrians. These methods only focus on a static image instead of dynamic video frames and therefore fall short for vision-based dynamic PID. Specifically, the dynamically changing AoIs and a number of pedestrians in video frames increase the difficulty and computational complexity of determining whether pedestrians intrude in the AoI. In addition, the lack of public datasets and benchmarks makes the vision-based dynamic PID methods difficult be evaluated.  

\begin{figure}[h]
  \includegraphics[width=\linewidth]{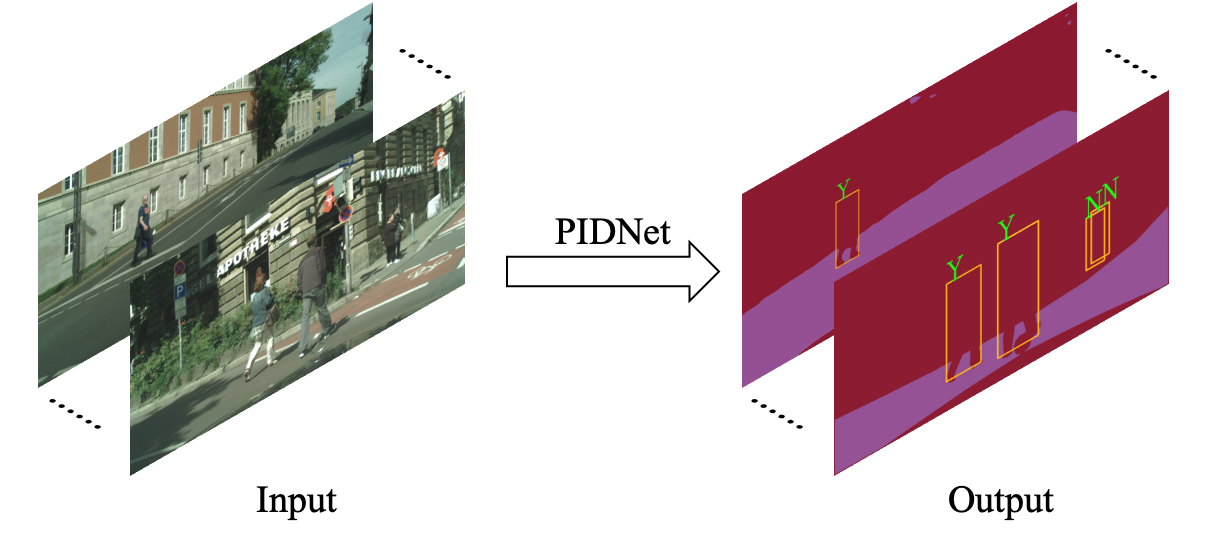}
  \caption{An illustration of our method on vision-based dynamic PID. The inputs are dynamically changing video frames and outputs are intrusion detection results of PIDNet. Letter ``Y" on the bounding box of the detected pedestrian indicates that the pedestrian has intruded into the AoI ( the AoI refers to the road in this paper unless otherwise specified), while the letter ``N" shows that the pedestrian has not intruded into the AoI.}
  \label{fig:intro}
\end{figure}

Recently, many deep-learning-based object detection \cite{yolo9000,fasterrcnn,ssd,maskrcnn} and semantic segmentation \cite{FCN,segnet,pspnet,bisenet} works have made great progress in dynamic scenarios analysis. Inspired by these works, we propose an efficient multi-task deep network, PIDNet, to solve the vision-based dynamic PID problem. PIDNet mainly consists of a semantic segmentation branch to accurately segment the dynamically changing AoIs and an object detection branch to quickly detect the intruded pedestrians. The final intrusion results are given by analyzing the pixel masking area between a detected pedestrian bounding box and a segmented AoI. Besides, to reduce the proposed network computational complexity and improve the network inference speed, three efficient network designs are developed: 1) a PID task feature extraction backbone is designed to share feature extraction of the two branches, 2) a feature cropping module is proposed to crop the feature map which only contains AoIs for quickly pedestrian detection use, and 3) a lighter detection branch is established by compressing the region proposal network and fully connected layers without affecting the accuracy. PIDNet in Figure \ref{fig:intro} can accurately localize the boundaries of varying forms of AoIs and pedestrians from dynamically changing video frames and then make intrusion judgment efficiently.

Noticing that we do not perform segmentation of AoIs and pedestrians simultaneously. Instead, we use a pedestrian’s bounding box to conduct the intrusion analysis. There are two reasons. The first one is to reduce the cost of labeling pedestrians. Compared with pedestrian bounding box annotation for detection purposes, pixel-level fine segmentation annotation for multiple pedestrians in an image is more time-consuming. Another is to reduce the computational cost of segmenting AoIs and pedestrians. Usually, there are multiple pedestrians in an image. Segmenting them requires a more complex network than that for segmenting AoIs only since the former needs to learn more parameters to segment pedestrians of different appearances and poses.

Considering the lack of public datasets and benchmarks, we create a vision-based dynamic PID dataset, Cityintrusion, and define corresponding evaluation metrics. Since vision-based dynamic PID needs both AoI segmentation and pedestrian detection labels in our method, we establish our dataset based on Cityscape \cite{cityscapes} (segmentation dataset) and Cityperson \cite{citypersons} (detection dataset). We conduct extensive experiments on the Cityintrusion dataset and the experimental results show that PIDNet can recognize most typical pedestrian intrusion cases, and have fast inference speed in vision-based dynamic PID task. In summary, our main contributions are listed as follows:

(1) A novel PID framework, PIDNet, which integrates AoIs segmentation and pedestrian detection together to perform pedestrian intrusion detection on dynamic video frames.

(2) Three efficient network designs, including feature sharing, cropping, and compression designs are developed to make the framework lighter and faster.

(3) A vision-based dynamic PID dataset, Cityintrusion, and the corresponding evaluation metrics are established to evaluate the proposed approach.

\section{RELATEDWORK}

\begin{figure*}[t]
  \includegraphics[width=\linewidth]{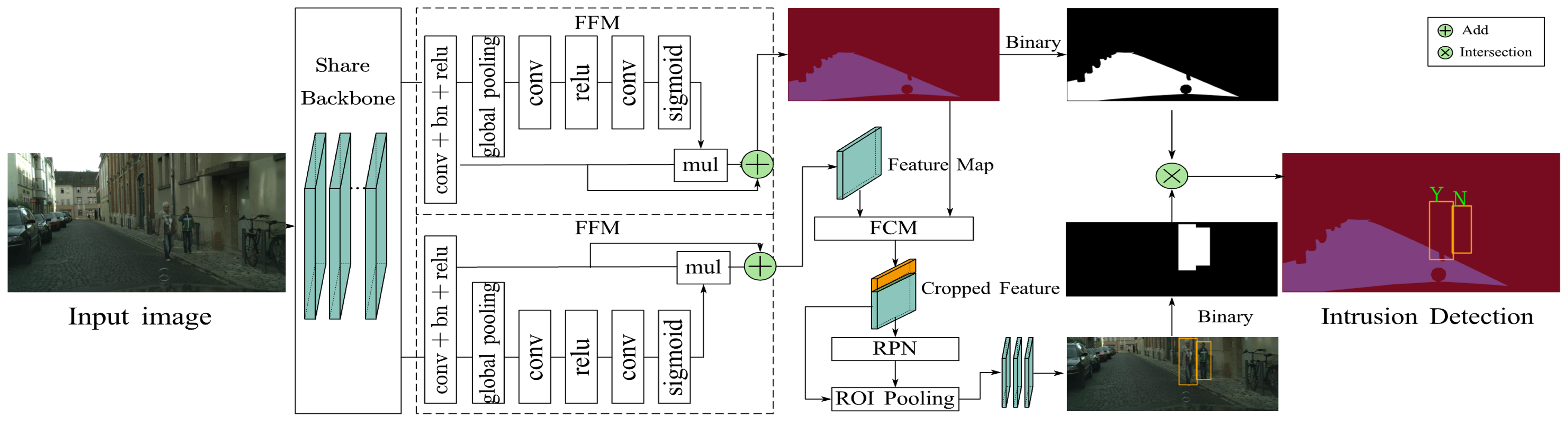}
  \caption{An illustration of the PIDNet structure. PIDNet mainly consists of a segmentation network in the upper branch and an object detection network in the lower branch. A special PID task feature extraction backbone is established to share feature for the two branches. Followed the backbone are two feature fusion modules (FFM) in the upper and lower branch respectively. Then a feature cropping module (FCM) is designed to project the segmented area which contains AoIs to the detection branch for quickly detection use. The cropped feature map is further forwarded to the compressed region proposal network (RPN) and its subsequent modules to attain pedestrian detection results. Finally, the intrusion judgment is made based on the intersection of segmented binary AoI and pedestrian detection.}
  \label{fig:PIDNet}
\end{figure*}

\subsection{Vision-based PID}
Previous studies on vision-based PID mainly focus on static PID, and mostly use handcrafted features \cite{liang2012real,zhang2015perimeter,matern2013automated} based methods to solve this problem. Wang et al. \cite{video5} exploited the frame difference and background subtraction to detect intrusion behavior. Zhang et al. \cite{video2} proposed an algorithm based on GMM (Gaussian Mixture Model) + SVM (Support Vector Machine) to identify the image foreground detection and pedestrian classification, and improve the robustness of the PID algorithm in a fixed scene. Chen et al. \cite{chen2014moving} proposed a retina-enhanced approach to reduce the highlighting’s affection and give the intrusion judgment. G. et al. \cite{prabhakar2012} propose an integrated approach for the tracking of abandoned objects using background subtraction and morphological filtering. However, the existing algorithms are tailored for fixed AoIs, which cannot be generalized to dynamic changing AoIs. 

\subsection{Dynamic vision Analysis}
Recently, many works have been proposed to address semantic segmentation \cite{FCN,segnet,u-net} or object detection \cite{yolo9000,yolov3,ssd,fasterrcnn,r-fcn,maskrcnn} works in dynamic scenarios, promoted by the latest advance in deep learning. ICNet \cite{icnet} introduces the cascade feature fusion unit to quickly achieve high-quality pixel-wise segmentation. PSPNet \cite{pspnet} produces good quality results on the scene parsing task through pyramid pooling modules.BiSeNet \cite{bisenet} makes a good balance between speed and accuracy through a new feature fusion module to combine context information and spatial information. As the state-of-art performance network in the semantic segmentation, BiSeNet is chosen as the basic segmentation network in many vision tasks. While in object detection, one-stage networks like SSD \cite{ssd} encapsulates all computation in a single network and achieve a good prediction speed. YOLO \cite{yolov1} directly predicts the bounding boxes and associated class probabilities. Two-stage object detection networks like Faster RCNN \cite{fasterrcnn} designs a region propose network to extract candidate bounding boxes by anchors. With an intuitionistic and clear structure, a good balance in accuracy and speed, Faster RCNN  is adopted by many object detection networks later, such as Mask R-CNN \cite{maskrcnn}, R-FCN \cite{r-fcn}, etc. To solve the dynamic PID task, it’s feasible to simultaneously use a segmentation network and a detection network. However, the repeated feature extraction of the two networks and the time-consuming detection and segmentation operations cause high computational complexity. It's essential to design an efficient multi-task network to deal with vision-based dynamic PID tasks.

\subsection{Efficient Model Designs}
\textbf{For feature sharing}, image segmentation and object detection require different feature extraction capabilities of backbones. BiSeNet proposed a lightweight backbone containing context path and space path for semantic segmentation. DetNet \cite{detnet} and Tiny-DSOD \cite{tiny-dsod} designs backbones specifically for object detection. Therefore, simply applying existing backbones to the shared backbone in our multi-task PIDNet is not suitable, as the task to be solved is different, we design a new multi-task shared backbone for PIDNet. \textbf{For feature cropping}, to our knowledge, there are very few works that handle this problem previously. \textbf{For feature compression}, recently many lightweight frameworks \cite{squeezenet,shufflenet,light} have been proposed for faster computing speed under resource-constrained computing devices. MobileNet \cite{mobilenets} and Xception \cite{chollet2017xception} uses depthwise separable convolutions to build lightweight deep neural networks. Inspired by this, we design some network compression strategies to make the proposed network lighter and faster.

\subsection{Dataset}
As our work needs to perform both image segmentation and detection to make intrusion detection decision, we review the segmentation and detection datasets. INRIA \cite{dalal2005histograms} and ETH \cite{ess2008mobile} represent early efforts to collect pedestrian datasets. Currently, KITTI \cite{kitti}, Caltech \cite{dollar2011pedestrian} and Cityperson \cite{citypersons} are the established pedestrian detection benchmarks. However, all of these existing datasets do not have intrusion or no-intrusion labels, which cannot be straightly used for the PID task. The Cityscape \cite{cityscapes} dataset is a segmentation dataset similar to earlier scene understanding challenges like Pascal VOC \cite{pascalvoc} and Microsoft COCO \cite{microsoft}, which provides pixel-wise segmentation for a number of semantic object classes. We find that Cityscape and Cityperson datasets aiming at segmentation and detection tasks respectively share the same groundtruth image. Therefore, we build a PID dataset, Cityintrusion, based on Cityscape and Cityperson.

\vspace{0.5cm}
\section{PIDNET}
\subsection{PIDNet Overview}
Figure \ref{fig:PIDNet} shows the network structure of PIDNet. The PIDNet is designed with a shared backbone for PID task feature extraction followed by two branches: the upper one for segmenting AoIs from dynamically changing environments, and the lower one for detecting whether pedestrians intrude the AoIs. In the upper branch, followed by the backbone is a feature fusion module (FFM) \cite{bisenet} to learn more specific features for the segmentation task and the final AoI segmentation result is obtained by sampling the output feature maps of FFM. In the lower branch, we also introduce an FFM after the shared backbone to further learn more specific features for the detection task. Then after the detection FFM, a region proposal network (RPN) \cite{fasterrcnn} and RoI pooling \cite{fasterrcnn} module is employed, with one full connection layer for regression and classification. A feature cropping module (FCM) is designed to connect the two branches, mapping the segmented area to the detection branch, making the detection feature map only contain the AoIs. The cropped feature map is further forwarded to the compressed RPN and its subsequent modules, where the output is detected pedestrian bounding boxes. The final intrusion detection judgment is made by analyzing the pixel masking area between a binarized detected pedestrian bounding box and a binarized segmented AoI. Once the overlapping pixels of the bounding box and the AoI exceed a certain threshold, an instruction warning on the detected pedestrian bounding box will be added, as shown in Figure \ref{fig:PIDNet}.

\subsection{Feature Sharing Design}

\begin{figure}[h]
  \includegraphics[width=\linewidth]{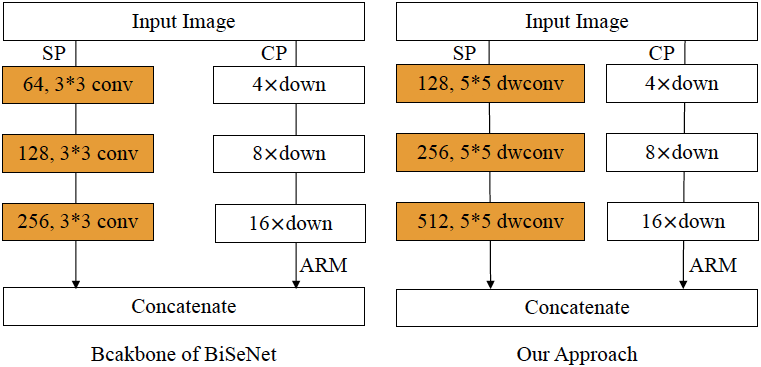}
  \caption{An illustration of our designed backbone and the backbone of BiSeNet. To extract more spatial information for dynamic PID task, we replace all 3*3 convolutions with 5*5 depthwise convolutions and double the channel numbers in the spatial path of BiSeNet.}
  \label{fig:backbone}
\end{figure}

The backbone is used to share feature extraction for the two branches. As the PID task requires special feature extraction capabilities (both segmentation and detection), simply applying existing segmentation or detection network backbone to our the multi-task PIDNet is not suitable, we hence establish a new backbone-BNet by modifying the backbone in BiSeNet. BiSeNet \cite{bisenet} uses a backbone that contains a spatial path (SP) to encode spatial information and a context path (CP) to extract context information. The SP contains three layers and each layer includes a convolution with a stride of 2, followed by batch normalization and ReLU. The CP utilizes a CNN model (usually resnet18 or resnet101 ) and global average pooling to provide a large receptive field. As mentioned in \cite{thundernet}, the context information is more important for the pixel-level segmentation task, while the spatial localization information is more crucial for the object detection task. The backbone in BiSeNet is focused more on the segmentation task and has limited spatial feature extraction capability. Therefore, we double the channel numbers of SP to achieve a good balance between spatial and context information preservation. Besides, as a large receptive field is essential both for segmentation and detection tasks, we replace all 3*3 convolutions in SP with 5*5 depthwise convolutions to enlarge the receptive field. This operation expands the receptive field without increasing the computational complexity. BNet and the original backbone of BiSeNet are shown in figure \ref{fig:backbone}.

\subsection{Feature Cropping Design}
\begin{figure}[h]
  \includegraphics[width=\linewidth]{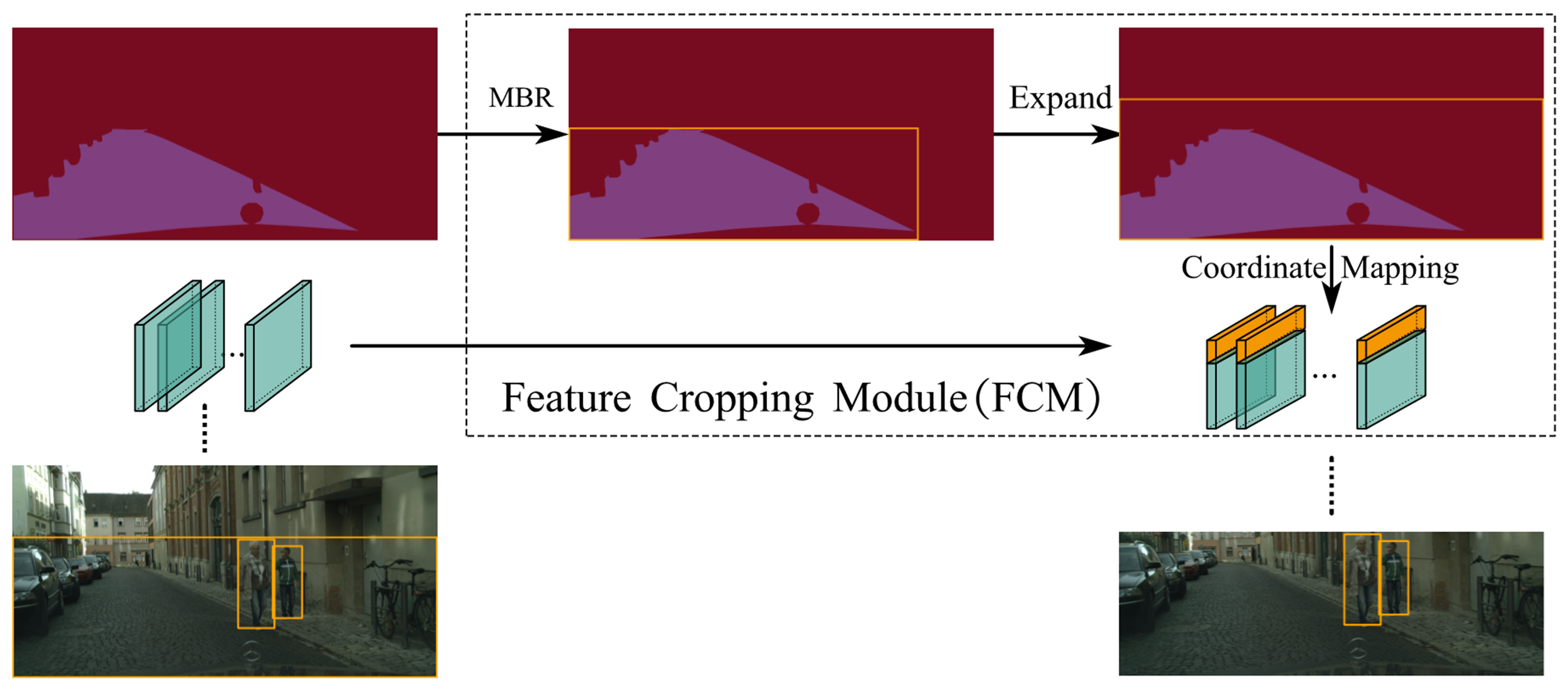}
  \caption{An illustration of the feature cropping module (FCM).  A minimum bounding rectangular (MBR) is generated around the segmented AoI. Then the expended rectangular is mapped to the whole feature map output by the detection branch to form a new compact feature map.}
  \label{fig:FCM}
\end{figure}

As the detection branch network needs to search the whole picture area (including AoI and no-AoI area) to detect pedestrians, which is time-consuming. We design a feature cropping module to remove the abundant feature information of the image that is not in the AoI, and project the segmented area to the feature map in the detection branch which only contains the AoIs for quickly detection use. As shown in Figure \ref{fig:FCM}, we generate a minimum bounding rectangular (MBR) around the segmented AoI from segmentation branch, and make slight extension to the rectangle to avoid miss errors. We then map this rectangle to the output feature map by the detection branch of the PIDNet to form a new compact feature map containing the AoI only.

Notice that the outer rectangle of the segmented contour miss body parts of the pedestrians (e.g., head, arm) on the road boundary. We design an extension strategy to avoid this as follows,
\begin{align}
\left\{\begin{matrix}
Y_{max}^{'}=\alpha (Y_{max}-Y_{min})+Y_{min}\\ 
X_{max}^{'}=\alpha (X_{max}-X_{min})+X_{min}
\end{matrix}\right.,
\label{eq1}
\end{align}
Where $\alpha$ stands for the extension coefficient of the rectangle, the value of $\alpha$ is determined as 1.2 experimentally. \!$(Y_{max},Y_{min},X_{max},X_{min})$\!, \!$(Y_{max}^{'},Y_{min},X_{max}^{'},X_{min})$\! are the coordinates before and after extension, respectively. If the rectangle coordinates exceed the original image size after extending, the coordinates of the image boundary are taken as the mapped rectangular coordinates. The feature mapping from the minimum external rectangle to the detection output feature map is 
\begin{align}
\left\{\begin{matrix}
y_{max}=\left \lfloor Y_{max}^{'}/s \right \rfloor+1,x_{max}=\left \lfloor X_{max}^{'}/s \right \rfloor+1\\
y_{min}=\left \lceil Y_{min}/s \right \rceil-1,x_{min}=\left \lceil X_{min}/s \right \rceil-1
\end{matrix}\right.,
\label{eq2}
\end{align}
where $(Y_{max}^{'},Y_{min},X_{max}^{'},X_{min})$ represent the rectangular coordinate of the segmentation image, $(y_{max},y_{min},x_{max},x_{min})$ denote the coordinates of the feature, and $s$ is the down-sampling times of the input image. The cropped feature map which contains the AoIs  area will be sent to the detection branch for further pedestrian detection. The process of feature cropping and mapping is also shown in Figure \ref{fig:FCM}. 

\subsection{Feature Compression Design}
Faster R-CNN is a representative two-stage object detection network used by many vision tasks, however, it cannot be straightly adopted for pedestrian detection in dynamic PID for two reasons. The first one is the multiple scales and ratios of anchors in RPN are designed for general object detection, while pedestrians in dynamic PID usually have smaller scales and various shapes due to occlusion, pose and clutter changes. The second one is that Faster R-CNN employs two computation-intensive fully connection layers for per RoI classification and regression, which is very time-consuming especially when a large number of region proposals are utilized.

\begin{figure}[h]
  \includegraphics[width=\linewidth]{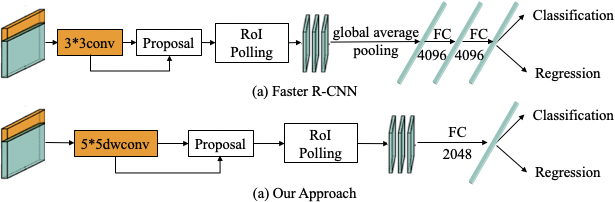}
  \caption{An illustration of feature compression strategy. In RPN, we replace the original 3*3 convolution with a 5*5 depthwise convolution and increase the number of aspect ratios and the anchor scales. In the R-CNN sub-network, we abandon the global average pooling and employ a single fully connection layer with 2048 channels.}
  \label{fig:rcnn}
\end{figure}

We establish our detection branch network by compressing the RPN and R-CNN sub-networks in Faster R-CNN \cite{fasterrcnn}. The compressed Faster R-CNN network structure is shown in Figure \ref{fig:rcnn}. To improve the accuracy and inference speed of PIDNet, we compress RPN by replacing the original 3*3 convolution with 5*5 depthwise convolution and 1*1 pointwise convolution to enlarge the receptive field. Besides, we set five different scales $\{32^2, 64^2, 128^2, 256^2, 512^2\}$ and five different aspect ratios $\{1:2, 2:3, 1:1, 3:2, 2:1\}$ of 25 anchors to cover pedestrians with different shapes and occlusion situations. Non-maximum suppression (NMS) \cite{fasterrcnn} is used to reduce the number of overlapping proposals. Moreover, to reduce PIDNet computational complexity, we only employ a single fully connection layer with 2048 channels with no dropout in the R-CNN sub-network, followed by are two fully connected layers respectively use for classification and regression. The original global average pooling followed by RoI Pooling operation, which is used to reduce the computation cost of the first fully connected layer, is also abandoned so that more spacial information is kept. 

\section{CITYINTRUSION}
\subsection{Dataset Annotation}
\begin{figure}[ht]
  \includegraphics[width=\linewidth]{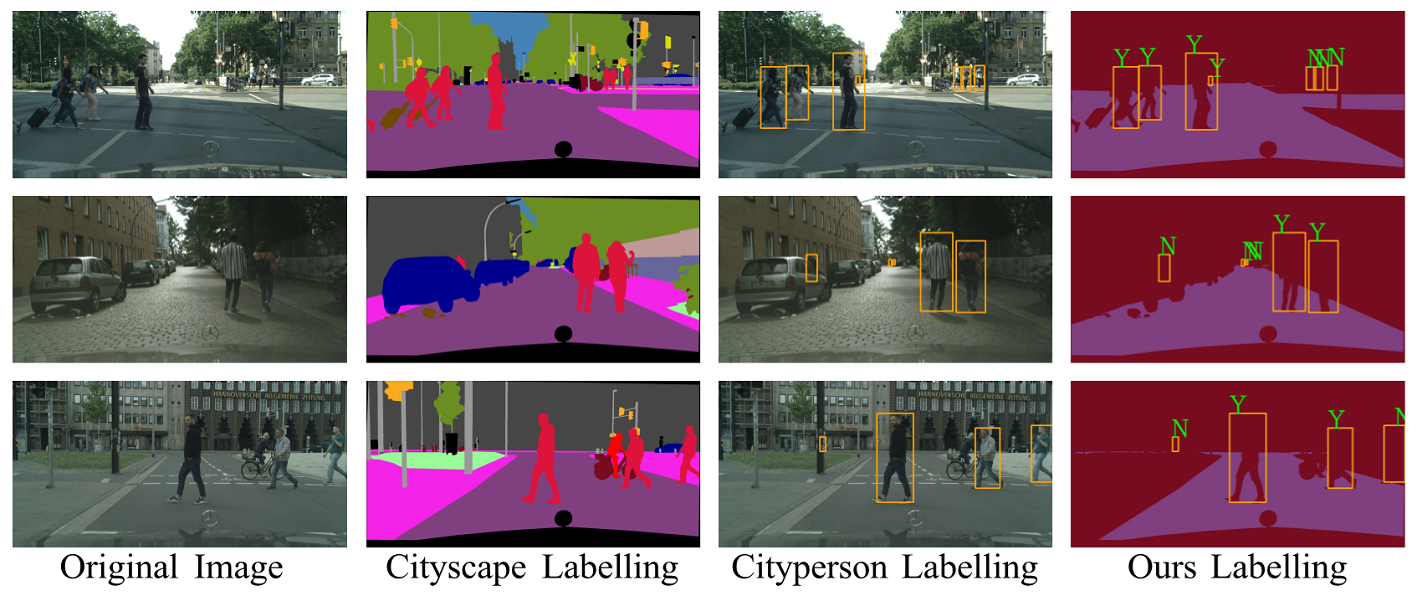}
  \caption{The illustration of labelling differences among Cityscape, Cityperson, and our Cityintrusion dataset.}
  \label{fig:dataset}
\end{figure}

To evaluate the proposed PIDNet, we create a dynamic PID dataset-Cityintrusion. An annotation team of 10 students is formed to construct, label and double-check the dataset manually. We establish our dataset based on Cityscape and Cityperson datasets and make modifications as follows. First, we remove a small amount of data that does not contain pedestrians in Cityperson dataset. Then we calculate overlapping pixel numbers between pedestrians bounding box and corresponding AoI (road here) with the help of instance segmentation labels in Cityscape dataset to give a preliminary label as “Intrusion” or “No Intrusion” for each person. At last, we manually calibrate the data and check the wrong labels to ensure the correctness of every labeled case. Figure \ref{fig:dataset} shows the labeling differences between three datasets for different purposes.  

\subsection{Dataset Statistics}

\begin{figure*}[ht]
\includegraphics[width=\linewidth]{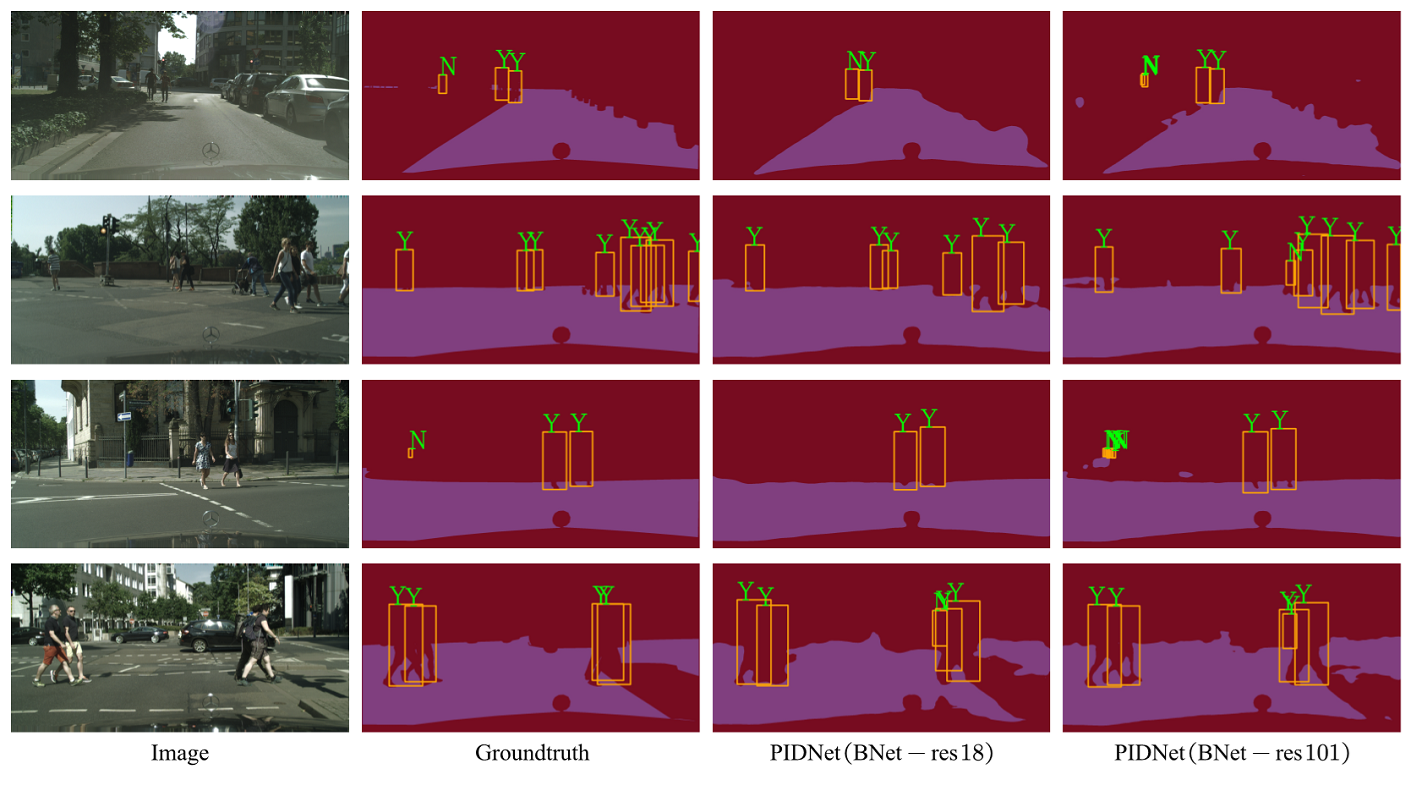}
\caption{Evaluation results on Cityintrusion dataset. The first column represents the real PID scenario, the second column represents the ground-truth of the dataset, and the third and fourth columns represent our prediction results. Our prediction results show that PIDNet can recognize most typical pedestrian intrusion cases, and the performance of PIDNet (BNet-res101) is better than PIDNet (BNet-res18).}
\label{fig:Final results}
\end{figure*}

The Cityintrusion dataset contains 2701 images with 4593 intrusion cases and 15078 non-intrusion cases. The statistics of Cityintrusion annotations are shown in the table \ref{tab:dataset}. The dataset is collected from 18 cities, covering 3 different seasons and various weather conditions, has on average ~7.3 intrusion and no-intrusion cases per image with each one considering real city road-based PID scenarios as much as possible. Since the test dataset of Cityscape is not publicly available, we set up the train and validation sets, and conduct experiments using the two sets. 

\begin{table}[h]
\begin{center}
\begin{tabular}{lccc}
\hline
Categories         & Train & Val  & Total \\ \hline
Cities              & 18    & 3    & 21    \\
Images             & 2303  & 398  & 2701  \\
Intrusion Cases    & 3829  & 770  & 4599  \\
No-Intrusion Cases & 12691 & 2393 & 15084 \\
Average cases per image    & 7.2   & 7.9  & 7.3   \\\hline
\end{tabular}

\caption{Statistics of Cityintrusion dataset. The intrusion case indicates that a person exists in the road, the average cases per image indicate the total number of intrusion cases and no-intrusion cases in one image.}
\label{tab:dataset}
\end{center}
\vspace{-0.7cm}
\end{table}

\subsection{Evaluation Metrics}
To evaluate the performance of the Cityintrusion dataset, we define two evaluation metrics, PID mean average precision ($PID\_mAP$) and PID accuracy ($PID\_Acc$). $PID\_mAP$ is an evaluation indicator that reflects the statistics of both the PID recall and precision values. Similar to the average precision calculation in object detection \cite{pascalvoc}, we define the PID average precision ($PID\_AP$) as follows,
\begin{align}
PID\_AP=\frac{1}{N}\sum_{r\in\left \{ 0,0.1,0.2,...1 \right \}}{max}\; pre(c,p)|re(c,p)\geqslant r
\label{eq4}
\end{align}
with the sample number $N=11$, the recall $re(c,p) = tp/(tp + fn)$, and precision
$pr(c,p) = tp/(tp + fp)$ for a given confidence value $c$ and the overlapping pixels number $p$. The $tp$ is the number of true positives, $fp$ is the number of false positives, and $fn$ is the number of false negatives. The difference between $PID\_AP$ and object detection $AP$ is that our true positive selection is more strict than that in object detection, as the $tp$ can be formulated as follows,
\begin{align}
&tp = tp + 1, if(IoU> 0.5) \cap  (c> c_{t})\cap (p> p_{t})
\label{eq4}
\end{align}
where a pedestrian is determined as ``intrusion" if it satisfies following three conditions: (1) the intersection-over-union $IoU$ ratio of the detection bounding box and ground truth is greater than 0.5; (2) the confidence score $c$ is greater than a certain threshold $c_{t}$; (3) the overlapping pixels number $p$ between the bounding box and the AoI is greater than a given threshold $p_{t}$. 

The overlapping pixels number threshold $p_{t}$ can be set as 5, 10, 20, 50 pixels, etc. The smaller the value of $p_{t}$ is, the more strictly the judgments of intrusion detection are made. In this paper, we set $p_{t}=20$, by varying the confidence score, we can get different groups of recall and precision values, and then get the PID mean average precision ($PID\_mAP$) over these values. In addition, we set the $c_{t}=0.8$, and compute the accuracy of PID, $PID\_Acc = (tp+fn)/total$ to observe the detection results, where $total$ represents the total number of groundtruths. 

All experiments are conducted on a Tesla V100 GPU card under CUDA 9.0 and CUDNN V7.0. An alternative training strategy is used to train the multi-task network PIDNet. For the segmentation branch, we use the mini-batch stochastic gradient descent (SGD) \cite{backpropagation} with a base learning rate of $2.5e^{-2}$, the momentum of 0.9 and weight decay of $1e^{-4}$. For the detection branch, we use the same SGD optimizer with batch size of 1, initial learning rate of $1e^{-3}$, learning decay of 0.1 and weight decay of $5e^{-4}$.

\section{EXPERIMENTS}
\subsection{Results}

\begin{table*}[ht]
\begin{center}
\begin{tabular}{@{}lccccc@{}}
\toprule
Model                    & Backbone        & PID\_mAP       & PID\_Acc       & Speed           & Params(M)      \\ \midrule
(a) PSPNet\cite{pspnet}+Faster R-CNN\cite{fasterrcnn}+Masking  & res101+vgg16    & 29.8          & 57.4          & 0.09fps           & 202.8          \\
(b) ICNet\cite{icnet}+Faster R-CNN\cite{fasterrcnn}+Masking   & resnet50+vgg16  & 34.5          & 61.1          & 0.15fps            & 184.6          \\
(c) BiSeNet\cite{bisenet}+Faster R-CNN\cite{fasterrcnn}+Masking & resnet 18+vgg16 & 36.7          & 63.1          & 0.18fps            & 150.7          \\\midrule
(d) PIDNet               & BNet-res18      & \textbf{36.7} & \textbf{63.3} & \textbf{9.6fps} & \textbf{105.8} \\
(e) PIDNet               & BNet-res101     & \textbf{49.2} & \textbf{67.1} & \textbf{5.4fps} & \textbf{138.7} \\ \bottomrule
\end{tabular}
\end{center}
\caption{Evaluation results on Cityintrusion dataset. The rows a, b and c in the table represent results obtained using two existing networks, and rows d and e indicate results obtained using our network. $PID\_mAP$ and $PID\_Acc$ are the evaluation metrics as described in Section 4.3.}
\label{tab:result}
\end{table*}

\begin{table*}[h]
\begin{center}
\begin{tabular}{lcccc}
\hline
Model            & Backbone    & Seg IoU       & Det AP        & Params(M)      \\ \hline
(a) PSPNet\cite{pspnet}       & res101      & 97.4          & -             & 65.7           \\
(b) ICNet\cite{icnet}        & resnet50    & 96.5          & -             & 47.5           \\
(c) BISeNet\cite{bisenet}      & resnet 18   & 97.8          & -             & 13.6           \\ \hline
(e) SSA-CNN\cite{ssa-cnn}      & vgg16       & -             & 68.7          & 131.2           \\
(e) TLL\cite{song2018small}    & res50       & -             & 71.3          & 108.7          \\
(f) MFR-CNN\cite{citypersons} & res18       & -             & 71.4          & 137.1          \\ \hline
(g) PIDNet       & BNet-res18  & \textbf{98.0} & \textbf{72.8} & \textbf{105.8} \\
(h) PIDNet       & BNet-res101 & \textbf{98.5} & \textbf{75.6} & \textbf{138.7} \\ \hline
\end{tabular}
\end{center}
\caption{Evaluation results on the Cityperson and Cityscape datasets. The rows a, b and c in the table represent the existing segmentation networks, and rows d, e and f indicate the existing detection networks, while rows g and h are our PIDNet. Seg IoU represents the IoU of the Segmentation network, Det AP represents the AP of the Detection network.}
\label{tab:single result}
\end{table*}

We conduct extensive experiments on the Cityintrusion dataset. The image input resolution of the PIDNet is 512 * 1024, and the results are shown in Figure \ref{fig:Final results} and Table \ref{tab:result}. The shared feature backbone BNet adopts resnet18 and resnet101 \cite{resnet} as the base model respectively, and it is represented by BNet-res18 and BNet-res101 in the table. To demonstrate our effective model design strategies, we select several state-of-the-art segmentation and detection networks for comparison, including PSPNet \cite{pspnet}, ICNet \cite{icnet} and BiSeNet \cite{bisenet} which are integrated with the Faster R-CNN respectively, without any efficient model design strategies.

From Table \ref{tab:result}, we can see that PIDNet has achieved 63.3\% (BNet-res18) and 67.1\% (BNet-res101) accuracy of PID, which surpasses the highest accuracy of 63.1\% by using the two existing models (Faster R-CNN and BiSeNet). Figure \ref{fig:Final results} well demonstrate the intrusion detection effect of our PIDNet. From Figure \ref{fig:Final results} we can see that most of the intrusion or no-intrusion cases can be correctly identified. Besides, simply combining existing segmentation and detection networks, without carefully efficient model designs, makes the model's total parameter size larger than that of our efficient PIDNet (e.g., 202.8M, 184.6M, 150.7M, vs 105.8M). Further, it can be seen that by fusing the feature sharing, feature cropping, feature compression designs, PIDNet achieves an inference speed of 9.6 fps (BNet-res18) and 5.4 fps (BNet-res101) which is much higher than the speed of using the integration of BiSeNet and Faster R-CNN (0.18 fps). To summarise, our PIDNet is better than using the heuristic combination of an existing segmentation and a detection networks in both accuracy and inference speed.

As this is the first work on vision-based dynamic PID, it is difficult to find other suitable PID works for comparison. Nevertheless, we still compare the proposed approach with single segmentation or detection networks. We select several state-of-the-art segmentation networks (PSPNet, ICNet, BiSeNet) and detection networks (SSA-CNN \cite{ssa-cnn}, TLL \cite{song2018small},  MFR-CNN \cite{citypersons}), and compare them with our network on Cityscape and Cityperson datasets. The results are given in the table \ref{tab:single result}. The intersection over the union (IoU) is used as the evaluation metric of segmentation network, while the average precision (AP) is used as the evaluation metric of object detection. It can be seen from the table that our network can achieve the IoUs of 98.0\% (BNet-res18) and 98.5\% (BNet-res101) for a single segmentation task, achieve the state-of-the-art level of the segmentation networks, even surpass the origin BiSeNet due to the larger receptive field in the spatial path of our designed backbone. When dealing with a single detection task, our network can reach high APs of 72.8\% and 75.6\%, surpassing the highest detection average precision,  MFR-CNN (71.4\% AP), in the table. The reason can be attributed to our base feature extraction module which can extract rich spacial information, and the newly defined anchors of RPN. To summarize, our PIDNet is also competitive on a single detection or segmentation task. 

\begin{table*}[h]
\begin{center}
\begin{tabular}{@{}lcccc@{}}
\toprule
Backbone                & Seg IoU & Det AP & PID\_Acc & Params(M) \\ \midrule
SP+CP                   & 97.8    & 69.6   & 59.8    & 12.5      \\
SP+CP+5*5               & 98.0    & 71.5   & 61.1    & 14.1      \\
SP+CP+5*5Dw             & 98.0    & 71.5   & 61.1    & 11.7      \\
SP+CP+5*5Dw+Add channel & 98.0    & 72.8   & 63.3    & 12.2      \\ \bottomrule
\end{tabular}
\end{center}
\caption{Ablation study on the shared backbone. ``5*5" in the table means the 5*5 convolutions, ``5*5Dw" in the table refers to the 5 * 5 depthwise convolutions, and ``Add channel" in the table represents increasing the number of channels of the spatial path in the backbone.}
\label{tab:backbone}
\end{table*}

\begin{table*}[h]
\begin{center}
\begin{tabular}{ccccccc}
\hline
PID Net                   & FC                        & FC-Extension              & M-RPN                     & C-RCNN                    & PID\_Acc       & Speed(fps)           \\ \hline
\checkmark &                           &                           &                           &                           & 61.5          & 3.6          \\
\checkmark & \checkmark &                           &                           &                           & 60.7          & 6.4          \\
\checkmark &                           & \checkmark &                           &                           & 61.4          & 6.1          \\
\checkmark &                           &                           & \checkmark &                           & 63.3          & 2.9          \\
\checkmark &                           &                           &                           & \checkmark & 61.3          & 7.4          \\
\checkmark &                           & \checkmark & \checkmark & \checkmark & \textbf{63.3} & \textbf{9.6} \\ \hline
\end{tabular}
\end{center}
\caption{Ablation studies on feature cropping module and network compression. FCM in the table means feature cropping module, M-RPN refers to the compressed region proposal network, and C-RCNN refers to the compact R-CNN in the detection branch.}
\label{tab:FCM}
\end{table*}

\subsection{Ablation Study}
\textbf{Feature Sharing.} We use different backbones to conduct ablation experiments on the Cityintrusion dataset, and the results are shown in table \ref{tab:backbone}. For the convenience of the experiment, we select $PID\_Acc$ and Parameters to evaluate the performance of the backbone on the PID task. The origin backbone (SP + CP) in BiSeNet is selected as the baseline. Firstly, we replace all 3*3 convolutions in the spacial path of the backbone with 5*5 conventional convolution, this operation improves both the segmentation IoU and detection AP due to extension of the receptive field, but it also leads to the increase of parameter size from 12.5M to 14.1M of the backbone. Then, we further replace the 5*5 convolution with depthwise convolutions, and the parameter size of the spatial path decreases to about 1/9 of the original size, and the total parameter size of the shared backbone is reduced from 14.1M to 11.7M. Finally, we increase the channel numbers in the spatial path to encode more spatial Information. Although this operation increases the total parameter size from 11.7M to 12.2M, it improves the detection AP 71.5\% to 72.8\% and the PID accuracy from 61.1\% to 63.3\%. Overall, when we use all the strategies (5 * 5 depthwise convolution and add the channel numbers), the shared backbone achieves a 3.5\% accuracy improvement in the PID task while maintains the same memory consumption as the original network.

\textbf{Feature Cropping.} 
The ablation study on feature cropping strategy is shown in table \ref{tab:FCM}. The PIDNet with a feature sharing backbone is used as the baseline on the Cityintrusion dataset. The FCM represents feature cropping strategy while FCM-Extension represents feature cropping with an extension strategy. From table \ref{tab:FCM}, we can see that the feature cropping module improves the inference speed (from 3.6 fps to 6.4 fps) of the network. However, the cropped feature may cause some pedestrians around the boundary of rectangular to be separated apart, resulting in a decrease of PID accuracy decrease from 61.5\% to 60.7\%. Therefore, an extension strategy on the rectangular is applied to the feature cropping strategy (FCM-Extension). From the table, we can see that the extension operation improve the $PID Acc$ from 60.7\% to 61.4\% by contains more pedestrians on the boundary of the rectangular with a little decrease in inference speed from 6.4fps to 6.1fps. Compared with the baseline, the feature cropping and extension strategies significantly improve the original inference speed from 3.6fps to 6.1fps with a little impact on PID accuracy.

\textbf{Feature Compression.} 
The ablation studies on feature compression strategy are shown in table \ref{tab:FCM}. PIDNet with a feature shared backbone is selected as the baseline as well. We modify the RPN in the detection branch by replacing the convolution layers and increasing the number of anchor scales and aspect ratios. The modified RPN is represented by M-RPN in the table. From table \ref{tab:FCM} we can see that M-RPN makes the network inference speed slightly decrease, but improves the accuracy of PID from 61.5\% to 63.3\%. We also compress the R-CNN sub-network in the detection branch by employing a single fully connection layer with 2048 channels. This operation is represented by C-RCNN in table \ref{tab:FCM}. The experimental results show that  C-RCNN can improve the inference speed from 3.6fps to 7.4fps with a slight decrease in the accuracy of the network. If we use all the feature compression strategies (M-RPN and C-RCNN) and feature cropping and extension strategies, the $PID Acc$ improves from 61.5\% to 63.3\%, while the inference speed improves from 3.6 fps to 9.6 fps, both the accuracy and speed of the PIDNet are improved significantly, which demonstrate that our designs are efficient.

\section{CONCLUSION}
This work proposes an efficient network, PIDNet, which fuses an AoIs segmentation network and a pedestrian detection network together and make deep improvements according to the dynamic PID task-specific requirement, solves the difficulty of pedestrian intrusion detection on a moving camera successfully. Besides, three efficient network design strategies including feature sharing, cropping, and compression are proposed and incorporated into the PIDNet to make it lighter and faster. Finally, a vision-based dynamic PID dataset and the corresponding evaluation metrics are established for PID research. Experimental results show that PIDNet can recognize most typical pedestrian intrusion cases accurately and quickly, and the results serve as a good baseline for the future vision-based dynamic PID study.

\begin{acks}
This work is supported by National Natural Science Foundation of China (No. 61790571), Shanghai Pujiang Program (No.19PJ1402000), Key Area R\&D Program of Guangdong Province (No. 2018B030338001), Consulting Research Project of the Chinese Academy of Engineering (No. 2019-XZ-7), Shanghai Engineering Research Center of AI \& Robotics, Engineering Research Center of AI \& Robotics, Ministry of Education 
in China and  Alibaba-Zhejiang University Joint Research Institute of Frontier Technologies.
\end{acks}

\balance
\bibliographystyle{ACM-Reference-Format}
\bibliography{sample-base}

\end{document}